%% file: acl_latex.tex
\pdfoutput=1

\documentclass[11pt]{article}

\usepackage[]{acl}

\usepackage{times}
\usepackage{latexsym}
\usepackage{amsmath}

\usepackage{amsmath}
\usepackage{graphicx, array, blindtext}
\usepackage{multirow}
\usepackage{hhline}
\usepackage{subcaption}
\usepackage{amsfonts,amssymb}
\usepackage{textcomp}
\usepackage{gensymb}
\usepackage{enumitem}
\usepackage{comment}
\usepackage{caption}
\usepackage[switch]{lineno}

\usepackage{pifont}
\newcommand{\cmark}{\ding{51}}
\newcommand{\xmark}{\ding{55}}
\newcommand{\greencheck}{\color{green!70!brown}{\cmark}}
\newcommand{\redx}{\color{red}{\xmark}}

\usepackage[T1]{fontenc}

\usepackage[utf8]{inputenc}

\usepackage{microtype}

\title{On the Limits of Evaluating Embodied Agent Model Generalization\\Using Validation Sets}

\author{Hyounghun Kim\textsuperscript{\rm 1}\;
Aishwarya Padmakumar\textsuperscript{\rm 2}\;
Di Jin\textsuperscript{\rm 2}\\
\textbf{Mohit Bansal\textsuperscript{\rm 1,\rm 2}}\;
\textbf{Dilek Hakkani-Tur\textsuperscript{\rm 2}} \\
\textsuperscript{1}UNC Chapel Hill \;\;\;
\textsuperscript{2}Amazon Alexa AI \\
\{hyounghk, mbansal\}@cs.unc.edu  \;\;\;
\{padmakua, djinamzn, hakkanit\}@amazon.com}

\newcommand{\modelname}{\textsc{VAM}}

\begin{document}
\maketitle
\begin{abstract}
Natural language guided embodied task completion is a challenging problem since it requires understanding natural language instructions, aligning them with egocentric visual observations, and choosing appropriate actions to execute in the environment to produce desired changes.
We experiment with augmenting a transformer model for this task with modules that effectively utilize a wider field of view and learn to choose whether the next step requires a navigation or manipulation action. 
We observed that the proposed modules resulted in improved, and in fact state-of-the-art performance on an unseen validation set of a popular benchmark dataset, ALFRED.  
However, our best model selected using the unseen validation set underperforms on the unseen test split of ALFRED, indicating that performance on the unseen validation set may not in itself be a sufficient indicator of whether model improvements generalize to unseen test sets.
We highlight this result as we believe it may be a wider phenomenon in machine learning tasks but primarily noticeable only in benchmarks that limit evaluations on test splits, and highlights the need to modify benchmark design to better account for variance in model performance.
\end{abstract}

\input{intro}

\input{dataset}
\input{model}
\input{expriment}
\input{results}

\input{conclusion}

\section*{Acknowledgments}
We thank the reviewers for their helpful comments. This work was partially done while Hyounghun Kim was interning at Amazon Alexa AI and later extended at UNC, where it
was supported by NSF Award 1840131 and DARPA KAIROS Grant FA8750-19-2-1004. The views contained in this article are those of the authors and not of the funding agency. 

\bibliography{anthology,custom}
\bibliographystyle{acl_natbib}

\end{document}

%% file: intro.tex
\section{Introduction}
Language guided embodied task completion is an important skill for embodied agents requiring them to follow natural language instructions to navigate in their environment and manipulate objects to complete tasks.
Natural language is an easy medium for users to interact with embodied agents and effective use of natural language instructions can enable agents to navigate more easily in previously unexplored environments, and complete tasks involving novel combinations of object manipulations.
Vision and language navigation benchmarks~\cite{anderson2018room2room, thomason:corl19, ku2020room} provide an agent with natural language route instructions and evaluate their ability to follow these to navigate to a target location. It requires agents to have a deep understanding of natural language instructions, ground these in egocentric image observations and predict a sequence of actions in the environment.
Other benchmarks study the manipulation and arrangement of objects~\cite{bisk2016towards, wang2016learning, li2016spatial, bisk2018learning} - another crucial skill to complete many tasks that users may desire embodied agents to be able to complete. These tasks additionally require agents to reason about the states of objects and relations between them. 
Language guided embodied task completion benchmarks~\cite{ALFRED20, kim2020arramon, padmakumar2021teach} combine these skills -- requiring agents to perform both navigation and object manipulation/arrangement following natural language instructions.

In this paper, we explore a challenging navigation and manipulation benchmark, ALFRED~\cite{ALFRED20}, where an agent has to learn to follow complex hierarchical natural language instructions to complete tasks by navigating in a virtual environment and manipulating objects to produce desired state changes.  
The ALFRED benchmark provides a training dataset of action trajectories taken by an embodied agent in a variety of simulated indoor rooms paired with hierarchical natural language instructions describing the task to be accomplished and the steps to be taken to do so. For validation and testing of models, there are two splits each - seen and unseen splits. 
The seen validation and testing splits consist of instructions set in the same rooms as those in the training set, while the unseen splits consist of instructions set in rooms the agent has never seen before, with rooms in the unseen test set being different from those in the train and unseen validation set. 
Performance on the unseen validation and test sets are considered to be the best indicators of whether a model can really solve the task as the agent must operate in a completely novel floorplan, and cannot rely for example on memorized locations of large objects such as a fridge or a sink. 
Additionally, the ground truth action sequences are not publicly available for the seen and unseen test sets, and participants must submit prediction acted sequences on the test sets to an evaluation server where they are privately evaluated to obtain test performance. 
The evaluation server limits the number of submissions that can be made from an account to one per week to discourage directly tuning hyperparameters of a model on the test set. 
It is expected that following standard procedure in training machine learning models, one may use the validation sets to evaluate models trained with different hyperparameters, or ablating different components on the validation sets and only evaluate the best model on the test sets.
Since ideally we would want a model to perform well on the unseen test set, it is reasonable to use success rate on the unseen validation set as a metric to choose which model is to be submitted for evaluation on the unseen test set. 

One technique previously demonstrated to improve performance on ALFRED is the use of a multi-view setup~\cite{nguyen2021look, kim2021agent} where an agent turns or moves its head in place at every time step to obtain additional views before deciding what action to take.
In contrast to current models that simply concatenate features from each view, we use view-action matching - explicitly aligning embeddings of actions with embeddings of corresponding views - and using a score from fusing these aligned embeddings to select the next action to be taken.
 This is inspired by a dominant paradigm for modeling visual navigation tasks called viewpoint selection~\cite{fried2018speaker} where an agent predicts the next action by examining the resultant views each of those would produce and selecting the desired future view. Viewpoint selection is possible in some simulators such as R2R where the environment does not get altered by the agent's actions and the agent's movement is confined to a fixed grid. The ALFRED dataset uses the AI2-THOR simulator which supports a wider action space, physics modeling for movement and a more dynamic environment including irreversible actions. Hence, it is not possible to obtain the view that would result from an action without taking it, preventing direct application of viewpoint selection. 
Additionally, the agent must decide at each time step whether to perform navigation or manipulation actions. 
In contrast to prior work that uses a single classifier layer over all possible actions treating them equally, we propose a gate module which gives a higher weight to actions of a more relevant action type. 

We follow standard experimental procedure training our modified models on the train split and using success rate on the unseen validation split to compare to baselines and perform ablation studies. 
On this set, the proposed model equipped with the aforementioned modules outperforms the state-of-the-art multi-view setup approaches and the ablation study shows each proposed module helps improve the model's performance.

However, we observe an unexpected and large performance gap between the unseen validation and test data splits. 
Our model outperforms state-of-the-art baseline models on the unseen validation split, but performs worse than them on the unseen test split.
We hypothesize that it may be possible to overfit hyperparameters and design choices to one set of unseen environments (the unseen validation) and hence success on one such set of unseen environments is insufficient to guarantee that a model will generalize to another set of unseen environments (the unseen test).
We report this finding as we believe this situation is likely more common during development on machine learning benchmarks, but such intermediate results are unlikely to be published. Instead after a poor result on a test set, it is likely that researchers continue further model modifications until a model setting is obtained that performs well on the test set. We believe that such models are likely overfitting to the test set of the benchmark and may not generalize well to a new test set.

%% file: dataset.tex
\section{Dataset \& Environment}
In this paper, we focus on improving models for the ALFRED~\cite{ALFRED20} benchmark. ALFRED is built using the AI2-THOR simulator~\cite{ai2thor} which consists of 120 indoor scenes across 4 types of rooms. Scenes also contain a diverse set of objects that are rearranged in different configurations for each trajectory in the dataset. 
In ALFRED, a agent is given a high level natural language goal statement (\emph{``Put a chilled pan on the counter''}) as well as step by step natural language instructions corresponding to subgoals to be completed in order for achieving the goal (\emph{``Turn around and cross the room and then go right and turn to the left to face the stove ... Put the pan down on the counter to the right of the toaster''}).  
An agent has access to all these instructions at the start of the task and then has to iteratively predict navigation and manipulation actions in the environment based on egocentric image observations to complete subgoals in order.
An agent must predict between a discrete set of possible navigation and manipulation actions, and predict a segmentation mask for the object to be manipulated if a manipulation action is predicted.
The performance for an agent is evaluated by comparing the final states of the objects at the end of the action trajectory executed by the agent to the states of the objects at the end of the ground truth trajectory.

%% file: model.tex
\section{Model}
We employ a vision-language transformer, LXMERT~\cite{tan2019lxmert} as the base architecture for our model. We encode the language input using a learned word embedding and transformer layer, and action history using a linear layer. 
Following~\citet{pashevich2021episodic}, we extract image features using a faster R-CNN~\cite{ren2015faster} pretrained on images from the AI2-THOR simulator, and average-pool features of regions into a single vector.
The visual and action features are first combined via a liner layer, and then fused with language features through a cross modal transformer layer.

\vspace{3pt}
\par
\noindent\textbf{View-Action Matching.}
We collect the multiple views (front, left, right, up, down) and go through the aforementioned process to obtain a feature $V_i$ from the cross modal transformer for each view, and compute its matching score $M_i$ with the corresponding action embedding $A_i$ using a feedforward network.

\vspace{3pt}
\par
\noindent\textbf{Action-Type Gate.}
We additionally learn a gate vector using a linear layer over features of all views at the current time step to better distinguish between navigation and non-navigation actions. This layer is trained to predict high weights for actions of the same type as the ground truth action and low weights otherwise. The predicted weights are multiplied pointwise with match scores $M_i$ and the action with the highest resultant score is selected.  
For example, if the ground truth action at a particular time step is \texttt{Move forward}, the gate will ensure that a prediction of \texttt{ToggleOff} which is a non-navigation action will receive a higher loss than a prediction of \texttt{Turn Right}, which is also an incorrect action but of the same type as the ground truth action (navigation). 

\paragraph{Loss.}
The model is trained via cross-entropy losses for action (teacher-forcing) and object type.

%% file: expriment.tex
\begin{table}[t]
    \begin{center}
    \resizebox{0.99\columnwidth}{!}{
    \begin{tabular}{c | c | c | c | c | c}
         &\multirow{2}{*}{Model} & Wide & View-Act & Act-Type & Success  \\
         && View & Matching & Gate & Rate (\%)  \\
         \hline
         1&Base LXMERT Architecture & \redx  & \redx & \redx & 4.7 \\
         2&\modelname{} (Ours) & \greencheck  & \redx & \redx & 9.3 \\
         3&\modelname{} (Ours) & \greencheck & \greencheck & \redx & 11.8\\
         4&\modelname{} (Ours) & \greencheck & \greencheck & \greencheck & 13.8 
    \end{tabular}
    }
    \caption{Performance improvement from wide view, view-action matching and action type gate modules on the ALFRED validation unseen split.}
    \label{tab:ablation}
    \end{center}
    \vspace{-10pt}
\end{table}
\section{Experiments}
\vspace{3pt}
\par
\noindent\textbf{Implementation \& Training Details.}
We use 2 language and 2 cross-modal LXMERT layers for the model, and use 768 as the hidden size. We use AdamW \cite{loshchilov2018decoupled} as the optimizer with the learning rate $1\times 10^{-5}$. All of the experiments are run on AWS `p3.16xlarge' EC2 instances running Ubuntu 18.04. We employ PyTorch \cite{paszke2017automatic} to build our models.

\vspace{3pt}
\par
\noindent\textbf{Data Splits.}
Following~\citet{ALFRED20}, we train our models on the train split and use success rate on the unseen validation split to perform model selection, and determine whether our model changes are likely to improve over existing state of the art models. 
We used the validation splits to evaluate the efficacy of variants of the transformer architecture, number of layers and number of epochs of training to use. 
We then submitted predictions from the best performing model on the unseen validation split to the evaluation server to obtain scores on the test sets.

\vspace{3pt}
\par
\noindent\textbf{Evaluation Metrics.}
We report two evaluation metrics from ~\citet{ALFRED20} on validation and test splits. Success rate (SR) measures the fraction of episodes whether the predicted model trajectory results in all object state changes produced by the ground truth action trajectory. Goal Condition Success Rate (GC) measures the fraction of such desired state changes across all episodes that were accomplished by model-predicted trajectories. 

\vspace{3pt}
\par
\noindent\textbf{Model Comparison.}
Recently, the best performing models on the ALFRED benchmark make use of semantic map representations of the environment~\cite{blukis2021persistent}. However, these rely on pre-exploration of the environment to build a semantic map, rather than utilizing language instructions to directly navigate to target objects.
Therefore, we focus on comparing our model with other multi-view setup models that are the state-of-the-art among non-SLAM models. LWIT~\cite{nguyen2021look} predicts an initial actions from an selected instruction alone and integrates the actions sequence with visual information to generate final actions to take. 
ABP~\cite{kim2021agent} factorizes the model into interactive perception and action policy modules for adapting to two different tasks (the former needs a pixel-level and the latter requires a global information). However, although they employ multi-view setup, the information from each view collapses into one integrated feature. On the other hand, our model exploit each view directly to keep the useful clues without any loss.

%% file: results.tex
\section{Results}

\begin{table}[t]
    \begin{center}
    \resizebox{0.95\columnwidth}{!}{
    \begin{tabular}{c | c | c | c }
         \multirow{2}{*}{Subgoals} & Wide & (+) View-Act & (+) Act-Type \\
         & View & Matching & Gate \\
         \hline
         \texttt{CleanObject} & 81.4 & 89.4 & 91.2\\
         \texttt{CoolObject} & 100.0 & 100.0 & 100.0\\
         \texttt{GotoLocation} & 62.0 & 66.2  & 67.1\\
         \texttt{HeatObject} & 100.0 & 100.0 & 98.5\\
         \texttt{PickupObject} & 69.2 & 68.5 & 68.5\\
         \texttt{PutObject} & 66.6 & 71.2 & 68.3 \\
         \texttt{SliceObject} & 62.2 & 61.3 & 69.4\\
         \texttt{ToggleObject} & 51.4 & 42.2 & 41.6\\
         
    \end{tabular}
    }
    \caption{Success rate (\%) of the sub-goal tasks on the ALFRED validation unseen split. }
    \vspace{-5pt}
    \label{tab:subgoal}
    \end{center}
    \vspace{-10pt}
\end{table}

\vspace{3pt}
\par
We first evaluate the utility of each modeling change on the unseen validation set of ALFRED. As shown Table~\ref{tab:ablation}, we gain 4.6\% on success rate from adding a wider field of view, an additional 2.5\% from view-action matching and a further 2\% from action type gating.
We observe a variance of ~3\% in success rate of the same type of model trained with different random seeds so we consider a 4.6\% improvement to be sufficiently large to be unlikely from pure variance. 

\vspace{3pt}
\par
\noindent\textbf{Sub-Goal Performance.}
Considering the proportion of \texttt{GotoLocation} to the total number of sub-goal tasks (i.e., 48\%) and its role of bridging other sub-goal tasks, navigation is very crucial ability for a agent to successfully perform this challenging ALFRED task. As shown in Table~\ref{tab:subgoal}, our full view-action matching (\modelname{}) model improves the performance of \texttt{GotoLocation} task by 5.1\% while also improves performance for some of other sub-goal tasks. This performance boost could attribute to the agent's ability to figure out where to go (View-Action Matching) and what to do (Action-Type Gate). 

\vspace{3pt}
\par
\noindent\textbf{Validation-Test Performance Gap.}
When we compare to other baselines in Table~\ref{tab:result}, although our model outperforms other state-of-the-art models on the unseen validation split by a large margin, its performance on the unseen test split is poorer, whereas the reverse trend is seen with ABP~\cite{kim2021agent}. 
This suggests that good performance from a model on an unseen validation set may not be a good method to determine whether model changes are likely to generalize to another unseen test set. 

This lack of generalization is more likely in current embodied learning tasks such as vision-and-language navigation or embodied task completion in comparison to other machine learning tasks due to the way unseen test sets are defined in embodied learning tasks. While ALFRED in particular does not introduce new object categories at test time, both validation and test unseen environments are visually different, by design from the training environment and from each other. When we compare models on the validation set, we hope that an increase in performance denotes a model that is more capable of generalizing to \textit{any} unseen environment. However, it may only be the case that the model only generalizes better to the particular visual differences present in the unseen validation environment. 

When the benchmark limits access to the test set, as in ALFRED, when dealing with a model that demonstrates variance when trained with different random seeds, hyperparameters and across training epochs, it is natural to choose the setting that results in the highest performance on the unseen validation set. However, a different setting may in fact be optimal for the unseen test set due to visual differences. While such a design is likely significantly more computationally expensive, it may be necessary to redesign benchmarks to take an average of performance from a few different variants of a model to reliably rank different modelling methods, instead of using scores from individual runs. We may also want to re-evaluate the value of keeping a test set private, as in the case of ALFRED that avoids prevents allowing models to overfit on the test set, but also makes it difficult to analyze the robustness of model performance between the validation and test sets. We would also like to encourage the reviewing community to enable the publication of modelling techniques whose performance is in the same ballpark as existing state-of-the-art models, but novel in some way, as opposed to solely relying on a model achieving a top score on a leaderboard as a criterion for publication, as this limits the development that could be made using these alternative modeling approaches.

%% file: conclusion.tex
\begin{table}[t]
    \begin{center}
    \resizebox{0.85\columnwidth}{!}{
    \begin{tabular}{c |c|c|c|c|c}
         \multirow{2}{*}{Split} & \multirow{2}{*}{Model} & \multicolumn{2}{c|}{Seen} & \multicolumn{2}{c}{Unseen}   \\
         \cline{3-6}
         &&SR&  GC &SR&  GC \\
         \hline
       \multirow{3}{*}{Val}  & LWIT &33.70 & 43.10 &9.70& 23.10\\
         &  ABP & 42.93 & 50.45 &12.55& 25.19 \\
         &  \modelname{} (Ours) &40.9 & 47.9&13.8& 28.1 \\
         \hline
        \multirow{3}{*}{Test}  & LWIT & 29.16 & 38.82 &8.37 & 19.13\\
         &  ABP &44.55 & 51.13 &15.43 & 24.76\\
         &  \modelname{} (Ours) & 35.42 & 43.98 & 8.57 & 20.69  
    \end{tabular}
    }
    \caption{Success rate (\%) on the ALFRED evaluation splits (GC: Goal-Condition). Our model outperforms the state-of-the-art multi-view setup models on validation splits but not test splits.}
    \label{tab:result}
    \end{center}
    \vspace{-5pt}
\end{table}
\section{Conclusion}
We attempted to improve a transformer model for embodied task completion by enabling it to effectively uses multiple views via view-action matching and action-type gating. 
Our view-action matching module computes a matching score between each a view and the embedding of the action used to generate it, and the gate module gives a higher weight to a more appropriate action type. 
While our model outperformed relevant baselines on the ALFRED unseen validation split, the trend was reversed on the unseen test split, suggesting that it may not be possible to over-utilize a validation split when making model selection choices so that the resultant model does not perform well on the test split.
We choose to publish this result as we believe this phenomenon is likely more common than reported with machine learning benchmarks, but only noticeable to researchers when working on a benchmark with limited access to the test set. 
We additionally hope that our work encourages the publication of promising modelling approaches that do not work as reliably as expected, so that these can act as a guide to researchers to better inform their future directions.